\documentclass[10pt,twocolumn,letterpaper]{article}

\usepackage{iccv}
\usepackage{times}
\usepackage{epsfig}
\usepackage{graphicx}
\usepackage{amsmath, bm}
\usepackage{amssymb}
\usepackage[linesnumbered, boxed, ruled]{algorithm2e}

\usepackage[pagebackref=true,breaklinks=true,letterpaper=true,colorlinks,bookmarks=false]{hyperref}

\usepackage[disable]{todonotes}  % uncomment this to disable comments/todos

\newcommand{\todoR}[1]{\todo[color=orange, prepend, caption={\textbf{Riccardo}}]{#1}}

\iccvfinalcopy % *** Uncomment this line for the final submission

\def\iccvPaperID{8609} % *** Enter the ICCV Paper ID here

\ificcvfinal\fi

\begin{document}

%%%%%%%%% TITLE
\title{Effective Real Image Editing with Accelerated Iterative Diffusion Inversion}

\author{Zhihong Pan \enspace Riccardo Gherardi \enspace Xiufeng Xie \enspace Stephen Huang\\
Oppo Mobile Telecommunications Corp.\\
2479 E Bayshore Rd, Palo Alto, CA, USA\\
\iffalse
{\tt\small firstauthor@i1.org}
\fi
\iffalse
\and
Second Author\\
Institution2\\
First line of institution2 address\\
{\tt\small secondauthor@i2.org}
\fi
}

\maketitle
\ificcvfinal\thispagestyle{empty}\fi

%%%%%%%%% ABSTRACT
\begin{abstract}
%\noindent Real image editing and manipulation using modern generative models like the Generative Adversarial Network (GAN) is very challenging for various reasons.  One critical obstacle is the inversion process to map a real image to its corresponding noise vector in the latent space so that the same image could be reconstructed, before the editing could be applied. For more recent Denoising Diffusion Implicit Models (DDIM),  the deterministic inversion process is not reliable due to linearization assumption in each inversion step.  When the reconstruction process diverges from the inversion process, image editing based on such inversions lead to fail cases.  Various works have tried to improve the inversion stability but they often come with significant trade-off in computational efficiency.  Here we propose an accelerated iterative diffusion inversion method that leads to significantly improved reconstruction accuracy without additional overhead in memory or computation.  Combining with a proposed blended guidance technique, we have demonstrated a large range of image editing tasks can then be effectively done without using classifier-free guidance in inversion.  More importantly, compared with other diffusion inversion based works, the proposed process is shown to be more robust in fast image editing where only 10-20 diffusion steps are used.
\noindent Despite all recent progress, it is still challenging to edit and manipulate natural images with modern generative models.  When using Generative Adversarial Network (GAN), one major hurdle is in the inversion process mapping a real image to its corresponding noise vector in the latent space, since it is necessary to be able to reconstruct an image to edit its contents.  Likewise for Denoising Diffusion Implicit Models (DDIM), the linearization assumption in each inversion step makes the whole deterministic inversion process unreliable.  Existing approaches that have tackled the problem of inversion stability often incur in significant trade-offs in computational efficiency.  In this work we propose an Accelerated Iterative Diffusion Inversion method, dubbed \emph{AIDI}, that significantly improves reconstruction accuracy with minimal additional overhead in space and time complexity.  By using a novel blended guidance technique, we show that effective  results can be obtained on a large range of image editing tasks without large classifier-free guidance in inversion.  Furthermore, when compared with other diffusion inversion based works, our proposed process is shown to be more robust for fast image editing in the 10 and 20 diffusion steps' regimes.
\end{abstract}

%either modify the unconditional textual embedding or introduce an auxiliary noise vector for improved or near-perfect reconstruction, 
%Inspired by the success of fixed-point iteration in backward Euler method, h

%%%%%%%%%%%%%%%%%%%%%%%%%%%%%%%%%%%%%%%%%%%%%%%%%%%%%%%%%%%%%%%%%%%%%%%%%
\section{Introduction}
\label{sec:intro}

\noindent Diffusion models are a class of generative models that learn to generate high-quality images
by iteratively applying a denoising process from a random noisy starting point.  They are capable
to achieve state-of-the-art (SOTA) image quality since the very early stage of
denoising diffusion probabilistic models (DDPM)~\cite{ho_nips_2020} and 
score-based generative modeling~\cite{song_iclr_2020}.
While the number of sampling steps required for high quality image generation was initially very large,
several follow-up studies~\cite{nichol_icml_2021, bao_arxiv_2022, lu_arxiv_2022, lu_arxiv_2022+} have reduced significantly the 
number of steps without degrading the image quality, making the widespread use of diffusion models possible.
In particular, denoising diffusion implicit models (DDIM)~\cite{song_iclr_2021} are widely used for their speed and flexibility in deterministic and stochastic generations.
Further reducing the number of sampling steps for both image generation or editing is still nevertheless an open research problem.

%The publicly available Stable Diffusion model~\cite{sd_github_2022} has demonstrated similar text-to-image capabilities and enabled many follow-up works included ours.
Diffusion models were initially designed for image generation; for this reason, their usefulness for real image
editing is limited without a proper inversion process, similar to the inversion challenge~\cite{xia_arxiv_2021}
faced by real image editing using a Generative Adversarial Network (GAN)~\cite{goodfellow_nips_2014}.
GAN inversion is limited by the reduced dimensionality in latent space; diffusion inversion is comparably less
restricted as the latent space has the same dimensionality as the original image.
Na\"ive one-step inversion step, \emph{i.e.} simply perturbing an image with random noise, was initially used for early image editing
works such as SDEdit~\cite{meng_arxiv_2021} and Blended Diffusion~\cite{avrahami_cvpr_2022}.
\iffalse
 SDEdit~\cite{meng_arxiv_2021} adopted a simple
one-step inversion method by perturbing images from another domain to use as
a partially inverted image surrogate.  Blended diffusion~\cite{avrahami_cvpr_2022} used similar inversion with partially injected noises, but it is applied
to pixels of non-editing area at each iteration step to maintain coherence with the edited area.
\fi
Later, an Euler method based inversion process that applies deterministic step-by-step noise injection was used
for image-to-image translation in DDIB~\cite{su_iclr_2022} and DiffusionCLIP~\cite{kim_cvpr_2022}.
However, as shown in the text-guided image
editing tests in Prompt-to-Prompt (PTP)~\cite{hertz_arxiv_2022}, such inversion
is not reliable for real image editing because the inversion often leads to failed reconstruction even when no editing is performed.
Follow-up works like null-text inversion (NTI)~\cite{mokady_arxiv_2022} and exact diffusion inversion (EDICT)~\cite{wallace_arxiv_2022}
have focused on improving the reconstruction accuracy by introducing auxiliary variables
like the learned null-text embedding in NTI, or processes like the coupled diffusion process
in EDICT.   The reconstruction accuracy improvements come however with regressions in computational complexity
of the inversion and/or the editing processes.

\begin{figure}[t]
 \begin{center}
     \includegraphics[width=\linewidth]{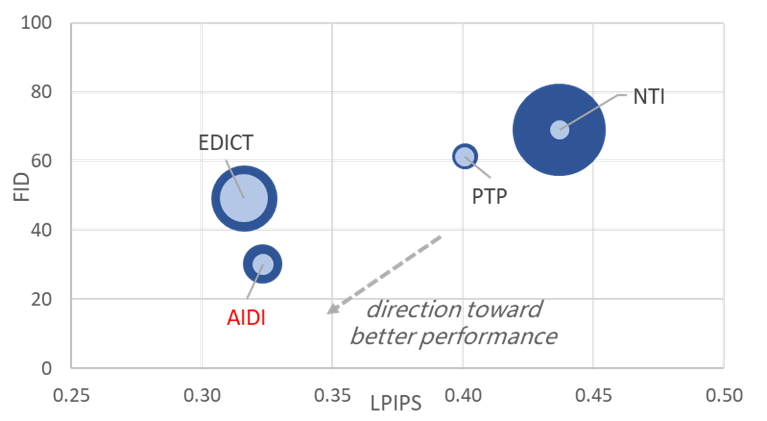}
 \end{center}
 \vspace{-6pt}
 \caption{Quantitative assessment for various diffusion inversion based methods, using a challenging image editing task of swapping dog to cat in AFHQ test set.  Image editing quality are assessed jointly by LPIPS and FID.  Lower LPIPS score is preferred for perception similarity with original image and lower FID in reference to the AFHQ cat train set translates to better image editing quality.  Circle areas and their outer ring represent average latency time for editing and inversion respectively.}
 \vspace{-9pt}
 \label{fig:first}
\end{figure}

In this paper, we are the first to look beyond the simple inversion process
that is based on the Euler method and investigate a better numerical solver
for improved inversion stability.  Modeling the inversion process
as an ordinary differential equation (ODE) problem, the implicit backward Euler method
is well suited as its solution results in an exact reconstruction using Euler's method, assuming the same time steps are used.
Given that, we propose an Accelerated Iterative Diffusion Iteration (AIDI)
method that adopts a fixed-point iteration process for each inversion step.
Combined with Anderson acceleration~\cite{anderson_jacm_1965} method which helps convergence of this iteration, it is demonstrated through
experiments of large test set that it results in significantly improved reconstruction accuracy.
Alternatively, an empirical acceleration method is invented for equivalent performance
with less computational overhead.

While a large classifier-free guidance~\cite{ho_nipsw_2021} scale is often needed for effective image editing, inversion with the same large guidance scale is not reliable even for our proposed AIDI.
We have demonstrated nevertheless that it is
possible to apply different guidance scales for inversion and editing respectively and
still achieve effective editing using our proposed blended guidance strategy.
Inspired by the cross-attention map replacement method
proposed in PTP~\cite{hertz_arxiv_2022}, we utilize the cross-attention map from the image
reconstruction using the same guidance setting of inversion to blend different guidance scales in image editing.
Higher guidance scales up to 7 are applied for pixels that require more editing and the low scale used in inversion, default as one,
is used for irrelevant pixels.

As shown in Fig.~\ref{fig:first}, using a challenging image editing task
that swaps dogs in high-resolution AFHQ~\cite{choi_cvpr_2020} images to cats, our method is the best overall in terms of both editing quality,
evaluated with the FID~\cite{heusel_nips_2017} score as related to the target cat domain in AFHQ, and perceptual similarity,
evaluated using the LPIPS~\cite{zhang_r_cvpr_2018} metric in reference to the input image.
Fig.~\ref{fig:first} also shows the average latency time as proportional circular areas.
Here the number of function evaluations (NFE) is not used because it is not an accurate metric of
computational complexity for all methods.
For example, in NTI there is substantial overhead in back propagation caused by learning the null-text embeddings in the inversion process.
Here we fix the number of diffusion steps to assess average latency time instead,
as it is correlated with image editing quality for all methods.
Note that a small number of 20 inversion and editing steps is used for all methods here.
As the inversion process is only needed once per image while editing could be applied multiple times, the latency time is split to editing and inversion, represented
as inner circle and outer ring respectively. 
For latency times, ours is only slower in inversion than the original PTP while as fast as PTP and NTI in editing.

In summary, based on pretrained text-to-image diffusion models, we propose a framework for  text-based real image editing with the following key advantages:
\begin{itemize}

\setlength\itemsep{0.01em}
\item[$\bullet$] We are the first to our knowledge to apply fixed-point iteration and acceleration in diffusion inversion, showing significant improvements in reconstruction accuracy
based on a large 5000 image COCO test set.  The LPIPS value for a 20-step reconstruction is reduced to 0.063 compared to 0.148 for the baseline.

\item[$\bullet$] For inversion without classifier-free guidance or low guidance scale, we propose a blended guidance method to apply larger guidance scales for effective editing in relevant areas, while maintaining fidelity elsewhere with low scales.

\item[$\bullet$] Our proposed image editing method is still effective for inversion steps as low 10, where competing approaches exhibit significant artifacts.

\end{itemize}

%%%%%%%%%%%%%%%%%%%%%%%%%%%%%%%%%%%%%%%%%%%%%%%%%%%%%%%%%%%%%%%%%%%%%%%%%
\section{Related Works}
\label{sec:rwork}

%-------------------------------------------------------------------------
\subsection{Text-to-Image Diffusion Models}
%\noindent\textbf{Text-to-Image Diffusion Models.}}
\noindent The rapid progress of Diffusion-based generative models has advanced the state-of-the-art for many generative tasks,
including text-to-image generations.
A highly capable unconditional diffusion model was shown in ~\cite{dd_github_2022}, using sampling guidance to match the CLIP scores
of the text input and generated image.  More recently techniques such as GLIDE\cite{nichol_arxiv_2022}, DALL$\cdot$E 2~\cite{ramesh_arxiv_2022} and Imagen\cite{saharia_arxiv_2022},
have used text embeddings from large language models to train conditional diffusion models,
all of them capable of generating diverse and high quality images that match with the arbitrarily complex prompt texts.
Both GLIDE and DALL$\cdot$E 2 are conditional to CLIP textual embeddings, while DALL$\cdot$E 2
trains instead a diffusion prior to first generate image embeddings from the input text CLIP embedding,
before the image embedding is then fed into another diffusion model for image generation.
To handle high-resolution image generation, both GLIDE and Imagen generate the text conditional image
at low-resolution using cascaded diffusion models, which are conditional to both text and image to
progressively increase resolution.  In alternative to that, LDM~\cite{rombach_cvpr_2022} proposed
to conduct the conditional text-to-image diffusion in a latent space of reduced dimensionality for faster training and sampling.
Based on the LDM architecture, a large
text-to-image model Stable Diffusion~\cite{sd_github_2022} was trained with a huge dataset and released for open
research.
\iffalse
Benefited from success of large scale language models like T5-XXL~\cite{raffel_jmlr_2020} and multi-model models like CLIP~\cite{radford_icml_2021},
DALL$\cdot$E 2~\cite{ramesh_arxiv_2022} and Imagen~\cite{saharia_arxiv_2022},
both large text-to-image diffusion models, have demonstrated outstanding ability of not only
generating high quality images that matches detailed instructive texts but also
creating realistic images from arbitrary descriptions.
\fi

%-------------------------------------------------------------------------
\subsection{Real Image Editing in Generative Models}
%\vspace{3pt} \noindent\textbf{{Real Image Editing.}}
%In recent years, GAN based deep learning models have been successful used for various
%generative tasks~\cite{donahue_arxiv_2018, tulyakov_cvpr_2018, karras_cvpr_2019},
%including text-to-image generations~\cite{reed_icml_2016, zhang_iccv_2017, xu_cvpr_2018, qiao_cvpr_2019, tao_arxiv_2020, frolov_nn_2021}.
\noindent Since the successful implementation of disentangled representation by StyleGAN~\cite{karras_cvpr_2019, karras_cvpr_2020}, image editing in GAN has become
very apt at separating different modes of variation such as pose, expression, color, and texture.
Various methods~\cite{patashnik_iccv_2021, gal_tog_2022} have been published for text-guided image editing using the contrastive language-image model CLIP.
While powerful for manipulating generated images, applying it to real image editing is not straightforward since
the inversion of a real image to latent variables is trivial.
\iffalse
There are two major type of GAN inversion methods: an optimization method that finds the optimal latent vector for a given
real image~\cite{abdal_iccv_2019, abdal_cvpr_2020, collins_cvpr_2020, kang_iccv_2021},
and a learning method that trains an encoder to learn the image to latent variable mapping using a large training
dataset~\cite{chai_iclr_2021, tov_tog_2021, richardson_cvpr_2021, alaluf_iccv_2021}.
\fi
Earlier GAN inversion methods focused on inversion without changing the GAN model weights, either via optimization~\cite{abdal_iccv_2019, abdal_cvpr_2020, collins_cvpr_2020, kang_iccv_2021}
or learning~\cite{chai_iclr_2021, tov_tog_2021, richardson_cvpr_2021, alaluf_iccv_2021}.
Two recent works, HyperStyle~\cite{alaluf_cvpr_2022} and HyperInverter~\cite{dinh_cvpr_2022},
introduced hypernetworks~\cite{ha_arxiv_2016} to modify the GAN weights for each image and help recovering lost
details during reconstruction.  It was shown that the modified weights have no adversarial effects on generated image quality when
the inverted latent variables are modified for image editing.  However, even with modified weights, GAN inversion is far from perfect due
to the significantly reduced dimensionality of the latent space as compared to the original image pixel space.

Diffusion models are not subject to this limitation since there is no dimensionality changes between input and output for each inversion step.
% However the works aiming to utilize the inversion process for optimal real image editing is still under continuous investigation and improving.
Earlier image editing methods using diffusion models, including SDEdit~\cite{meng_arxiv_2021} and blended diffusion~\cite{avrahami_cvpr_2022},
didn't utilize the potential of accurate diffusion inversion as they inject random noises into input image to synthesize a noisy start.
\iffalse
SDEdit~\cite{meng_arxiv_2021} was the first to use diffusion model for image translation and editing.
It injects noise to images of one domain, like a semantic map, and uses it as an equivalent partially inverted image before
it is used to regenerate a translated image in the target domain.  Later, blended diffusion~\cite{avrahami_cvpr_2022}
proposed local CLIP-guided diffusion to make realistic local image editing in a masked area.
To increase the coherence between masked and unmasked areas, similar partial noise injections are applied to unmasked pixels.
\fi
Another recent work Imagic~\cite{kawar_arxiv_2022} achieved mask-less real image editing without involving diffusion inversion, utilizing 
optimized textual embedding and diffusion model fine tuning instead.

\iffalse
The full-process inversion based on step-by-step noise injection using
the same generative diffusion model was first used for image-to-image translation in DDIB~\cite{su_iclr_2022},
which trains separate models for two different domains.
\fi
DiffusionCLIP~\cite{kim_cvpr_2022} was the first to adopt the more accurate step-by-step
diffusion inversion process but it relied on diffusion model refinement to achieve text-guided image editing, without addressing the inversion
accuracy challenge directly.  DiffEdit~\cite{couairon_arxiv_2022} also avoided the inaccuracy concern by controlling an encoding ratio and applying a generated mask.
Prompt-to-Prompt (PTP)~\cite{hertz_arxiv_2022} was the first to achieve comprehensive text guided image editing without diffusion model
refinement, including local editing without a known mask.  However it focused on generated image editing, citing that the simple step-by-step inversion is
not reliable for real images, especially with larger classifier-free guidance scales.
Null-text inversion (NTI)~\cite{mokady_arxiv_2022} proposed to change the constant null-text embedding to image-specific optimal ones in order to achieve accurate reconstruction and it then applies PTP techniques for real image editing.
Later EDICT~\cite{wallace_arxiv_2022} proposed to use an auxiliary diffusion branch to achieve exact inversion and reconstruction of the input image, resulting in improved image editing quality.
\iffalse
Direct Inversion~\cite{elarabawy_arxiv_2022} also stores the injected noise during inversion which are then mixed with the estimated noise during the editing process to preserve image similarity.
\fi
Most recently, pix2pix-zero~\cite{parmar_arxiv_2023} learns editing directions in the textual embedding space for corresponding
image translation tasks but it adopts the original diffusion inversion process used in PTP without making efforts to avoid the known
inversion divergence.  To our knowledge, our paper is the first to address the inversion accuracy challenge without change in model configuration or system architecture so the proposed iterative inversion could be applied to and benefit other methods like pix2pix-zero that do not address the inversion accuracy issue yet.

%%%%%%%%%%%%%%%%%%%%%%%%%%%%%%%%%%%%%%%%%%%%%%%%%%%%%%%%%%%%%%%%%%%%%%%%%
\section{Proposed Method}
\label{sec:method}

%-------------------------------------------------------------------------
\subsection{Diffusion Inversion Preliminaries}
\noindent Our proposed generative compression method is built on pre-trained, unmodified text-to-image diffusion models.
Without loss of generality, the publicly available Stable Diffusion model~\cite{sd_github_2022},
 which uses the latent diffusion model (LDM)~\cite{rombach_cvpr_2022} architecture, is adopted for all experiments.
 Given the LDM architecture, the diffusion process is conducted in the latent space $z$, which can be decoded to the image space $x$.
 Nevertheless our method is equally applicable to other diffusion models conducted in native image space instead of a latent space.

 For a text-to-image diffusion model learned from a large set of paired image latent variable $z$ and image caption $p$, the process is optimized
 using a simple standard mean-squared error (MSE) loss:
 \begin{equation}
 L^{\text{simple}} = E_{t, z, p, \epsilon} ||\epsilon - \bm{\epsilon}_{\theta}(z_t, p, t)||^2
 \label{eq:ls}
 \end{equation}
where $\epsilon$ is a random Gaussian noise added to $z$
to synthesize $z_t$, $t$ is a randomly-set time-step chosen from $1, 2, \ldots, T$, and $\bm{\epsilon}_\theta$ denotes the diffusion model trained to estimate the injected noise using optimized parameters $\theta$.
As the goal of our proposed iterative inversion is accurate reconstruction, here we use a DDIM sampler where the sampling noise is $\sigma_t = 0$ for all $t$ to achieve a deterministic sampling process.
The iterative sampling step to generate an image latent $z_0$ from a random sample $z_T$ is:
 \begin{equation}
 \begin{split}
    z^t_0 & = (z_t-\sqrt{1-\bar{\alpha}_t}\epsilon_t) / \sqrt{\bar{\alpha}_t} \\
    z_{t-1} & = \sqrt{\bar{\alpha}_{t-1}} z^t_0 + \sqrt{1-\bar{\alpha}_{t-1}} \epsilon_t
 \end{split}
 \label{eq:dn}
 \end{equation}
$\epsilon_t$ is typically calculated as follows:
\begin{equation}
    \epsilon_t = \omega \bm{\epsilon}_{\theta}(z_t, p, t) + (1-\omega) \bm{\epsilon}_{\theta}(z_t, \emptyset, t)
 \label{eq:eps}
 \end{equation}
where $\omega$ is the classifier-free guidance scale and $\emptyset$ is the null-text reference.

Early image editing works like SDEdit use the same noise injection step used in model training for inversion at any $t$.  That is:
\begin{equation}
    z_t = \sqrt{\bar{\alpha}_{t}} z_0 + \sqrt{1-\bar{\alpha}_{t}} \epsilon.
 \label{eq:s1}
 \end{equation}
Since $\epsilon$ is a random noise independent of $z_0$, this is a stochastic inversion process that cannot reliably reconstruct $z_0$
from $z_t$ when $t$ is close to $T$.  In later works,
a simple DDIM inversion process is adopted base on a linear ODE solver:
 \begin{equation}
 \begin{split}
    z^t_0 & = (z_t-\sqrt{1-\bar{\alpha}_t}\epsilon_{\Tilde{t}}) / \sqrt{\bar{\alpha}_t} \\
    z_{t+1} & = \sqrt{\bar{\alpha}_{t+1}} z^t_0 + \sqrt{1-\bar{\alpha}_{t+1}} \epsilon_{\Tilde{t}}.
 \end{split}
 \label{eq:df}
 \end{equation}
To allow for better reconstruction accuracy, here $\epsilon_{\Tilde{t}}$ is instead approximated using $t\!+\!1$ as follows:
 \begin{equation}
    \epsilon_{\Tilde{t}} = \omega \bm{\epsilon}_{\theta}(z_t, p, t+1) + (1-\omega) \bm{\epsilon}_{\theta}(z_t, \emptyset, t+1).
 \label{eq:epst}
 \end{equation}

\begin{figure}[t]
 \begin{center}
     \includegraphics[width=\linewidth]{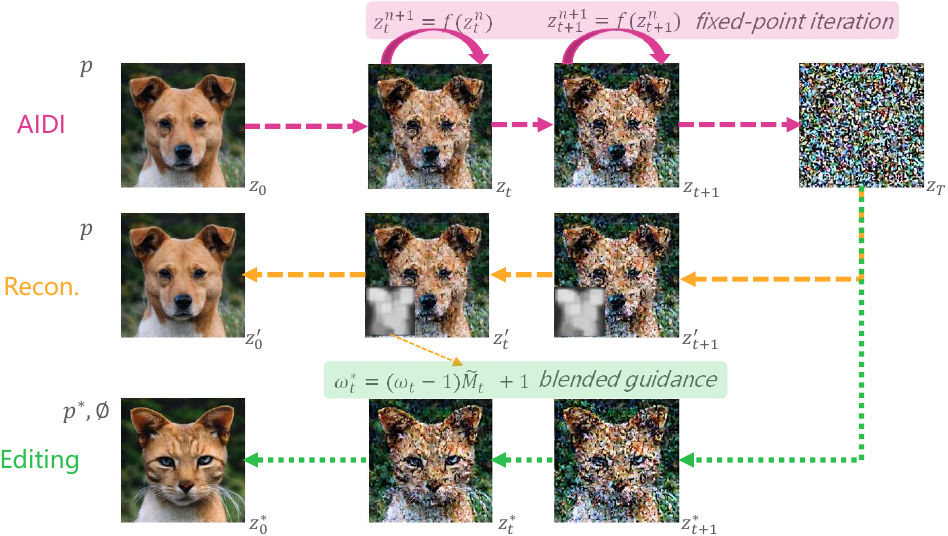}
 \end{center}
 \vspace{-6pt}
 \caption{Flowchart of the proposed effective real image editing. From top to bottom, (i) the image is transformed to an inverted noise vector using AIDI; the inverted noise vector is used to either (ii) reconstruct the original image using the same prompt $p$ or (iii) generate an edited image using prompts $p^*, \emptyset$ with classifier-free guidance, where the reconstruction process is also used to supply the mask for blended guidance and attention injection.  Note that the visible noise is not Gaussian, as all images are decoded from latent space $z$ to image space $x$ for display.}
 \vspace{-6pt}
 \label{fig:chart}
\end{figure}
\todoR{it would be nice to add a visual indication (number or letter) to the steps, to be able to refer to them in the caption}

%-------------------------------------------------------------------------
\subsection{Accelerated Iterative Diffusion Inversion}

\noindent For an inversion step where $z_{t-1}$ is known, we aim to find the optimal $z_t$ so that we can
recover $z_{t-1}$. We can rewrite Equation~\ref{eq:dn} as:
 \begin{equation}
    z_t \!=\! \sqrt{\frac{\bar{\alpha}_{t}}{\bar{\alpha}_{t-1}}} z_{t-1} \!+\! \left[ \sqrt{1-\bar{\alpha}_t} - \sqrt{\frac{(1-\bar{\alpha}_{t-1})\bar{\alpha}_{t}}{\bar{\alpha}_{t-1}}} \right] \epsilon_t.
 \label{eq:be}
 \end{equation}
Since $\epsilon_t$ depends on $z_t$ (\emph{cf.} Equation~\ref{eq:eps}), it can be denoted
as an implicit function $z_t = f(z_t)$. The ideal inversion step, finding $z_t$ that results in $z_{t-1}$ exactly, becomes a step to find a fixed-point solution for $f$.
In numerical analysis, this is often solved via the iterative process:
 \begin{equation}
    z^{n+1}_t = f(z^{n}_t), n = 0, 1, \ldots 
 \label{eq:it}
 \end{equation}

\begin{figure*}[t]
 \begin{center}
     \includegraphics[width=\linewidth]{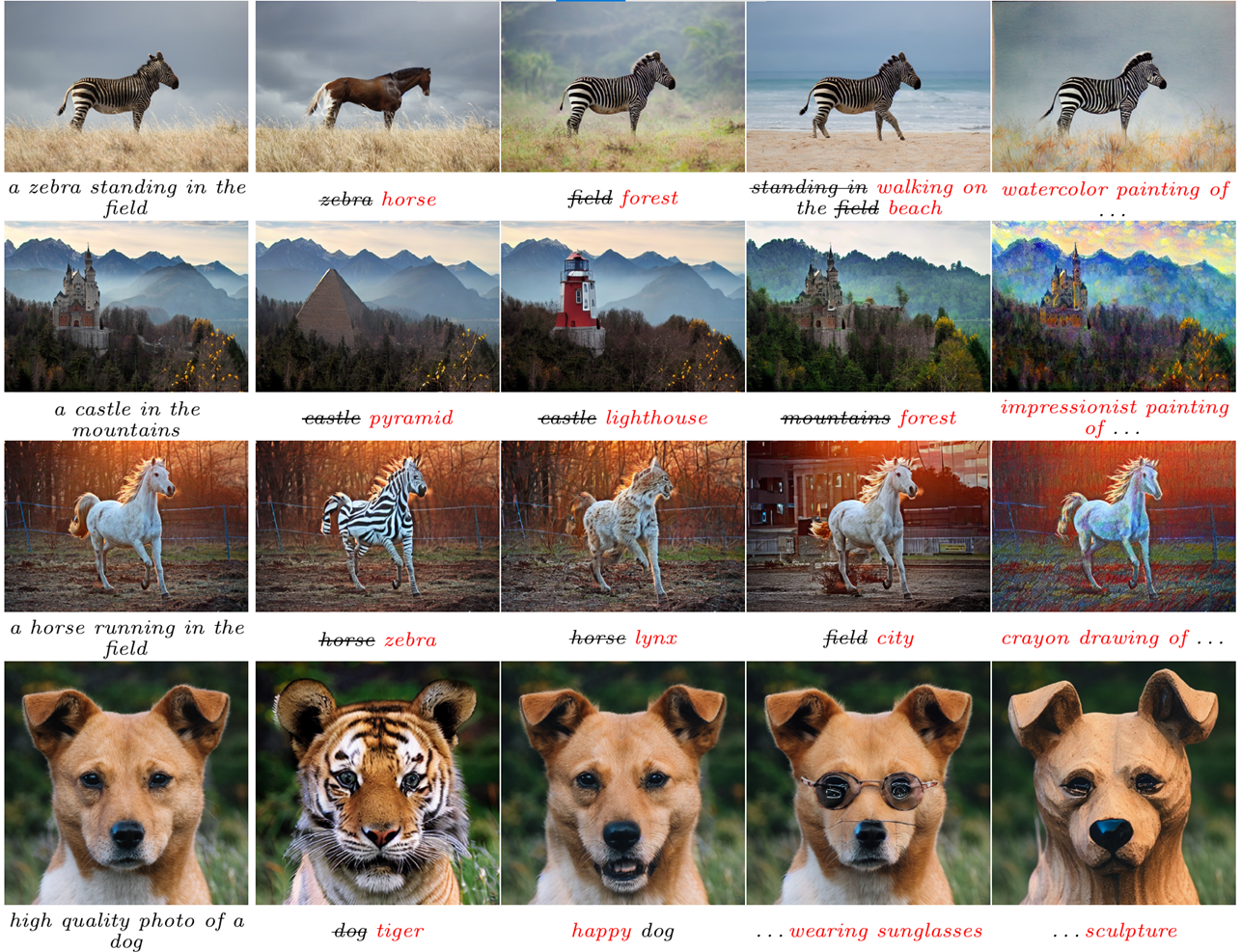}
 \end{center}
 \vspace{-6pt}
 \caption{Visual examples of the effective editing abilities of our proposed real image editing based on AIDI, using only 20 editing steps.}
 \vspace{-6pt}
 \label{fig:edit}
\end{figure*}
\todoR{This is awesome :)}

\begin{algorithm}[h]
\SetAlgoLined
\textbf{Input:} A latent image $z_0$ and prompt $p$, acceleration method $C$, iteration parameters $I, m$. \\
\textbf{Function}: $f(z)$ is the implicit function defined in Equation~\ref{eq:be}, $g(z)$ is the residual function $f(z)-z$\\
\textbf{Output:} An inverted noise vector $z_T$. \\
 \vspace{1mm} \hrule \vspace{1mm}
 \For{$t=1,2,\ldots,T$}{
    $z^0_t, z^1_t \leftarrow z_{t-1}, f(z^0_t)$ \;
    %$z^1_t \leftarrow f(z^0_t)$ \;
    \For{$i=1,\ldots,I$}{
        %$f_i \leftarrow f(z^i_t)$ \;
        \uIf {$C$ is $\mathrm{AIDI\_A}$} {
            $m_i \leftarrow min(m, i)$\;
            $G_i \leftarrow [g(z^{i-m_i}_t), \ldots, g(z^i_t)]$\;
            $\gamma_i \leftarrow {\arg\!\min}_{\gamma \in \Gamma_i} \|G_i \cdot \gamma \|_2$,  where $\Gamma_i = \{(\gamma^0, \ldots, \gamma^{m_i}): \sum_{j=0}^{m_i} \gamma^j = 1\}$\;
        }
        \uElseIf{$C$ is $\mathrm{AIDI\_E}$}{
            $m_i, \gamma_i \leftarrow 1, (0.5, 0.5)$\;
        }
        Set $z^{i+1}_t \leftarrow \sum_{j=0}^{m_i} \gamma^j_i f(z^{i-m_i+j}_t)$\;
   }
    $z_t \leftarrow z^I_t$ \;
 }
 \textbf{Return} $z_T$ 
 \caption{Accelerated Iterative Diffusion Inversion}
\label{alg:aidi}
\end{algorithm}

\noindent The convergence of this iterative process can often be accelerated using established techniques such as Anderson acceleration. We also propose and employ an alternative empirical acceleration method which is simpler and faster than Anderson's.  With appropriate acceleration,
the iterative inversion process can become more efficient and stabler than a simple forward Euler method. We summarize
our full proposed process in Alg.~\ref{alg:aidi}, and refer to it as \emph{AIDI}, short for accelerated iterative diffusion inversion.
Note that for the residual function $g(z)$ used in the AIDI\_A variant with Anderson acceleration, it is defined as $g(z)=f(z)-z$.
For the AIDI\_E variant, it is a simplified version of Anderson acceleration with fixed setting for $m$ and $\gamma$ as 1 and $(0.5, 0.5)$,
saving the additional optimization process to find $\gamma$.

%-------------------------------------------------------------------------
\subsection{Blended Guidance}
%It is pointed out in EDICT that higher order ODE solvers could improve inversion stability, there is no further investigate on its impact on reconstruction and subsequent editing quality for concerns about the trade-off between reconstruction accuracy and editing effect when large guidance scale is used.
\noindent While the proposed AIDI can significantly improve the inversion stability at different guidance scales, it is not sufficient on its own for reliably reconstructing and then eventually editing an image when the guidance scale is large.  Adopting the PTP pipeline
for real image editing, we first fix a small guidance scale like $\omega = 1$ for the inversion
using source prompt $p$.  For the following editing, we establish two parallel sampling processes, one conditional to prompt $p$ with same small $\omega$ and the other conditional to the target prompt $p^*$.
For the $p^*$ process, we introduce a blended $\omega^*$ to apply larger
guidance scales for pixels relevant to editing and lower ones for the rest
to keep them unedited.  To support this blended guidance, a soft editing mask $\Tilde{M}_t$ is generated concurrently with the image editing process.  First the cross-attention map $M_t$ with an anchor token is determined.  Here the anchor token refers to the word most relevant to the intended editing area.  This token could be positive, \emph{i.e.} associating with areas to edit, or negative, which associates it with areas to keep unedited.  In the case of $\textit{photo of a dog} \rightarrow \textit{photo of a cat}$, $\textit{dog}$ is a positive token.  In contrast, for $\textit{a dog} \rightarrow \textit{a dog on the beach}$, $\textit{dog}$ is a negative one instead.  The mask $M_t$ is first normalized to $\overline{M}_t$, where all pixels smaller than a threshold $\delta$ are normalized to the range of $(-M, 0)$ and the others normalized to the range of $(0, M)$.
Then a soft mask $\Tilde{M}_t$ is defined as $Sigmoid(\overline{M_t})$ for a positive token and $Sigmoid(-\overline{M_t})$ for a negative one.  Given that, the blended guidance scale $\omega^*$ is defined as
\begin{equation}
    \omega^*_t(k) = (\omega_E-\omega) \Tilde{M}_t(k) + \omega
 \label{eq:bgs}
 \end{equation}
where $\omega_E$ is a large guidance scale intended for editing, and $k$ refers to pixels of the mask image.    Note that the soft mask approaches a binary one when $M$ becomes very large.  
Combining our proposed AIDI and blended guidance, the overall process for our effective real image editing is illustrated in Fig.~\ref{fig:chart}, assuming $\omega=1$ in inversion for simplicity.

\begin{figure*}[t]
 \begin{center}
     \includegraphics[width=\linewidth]{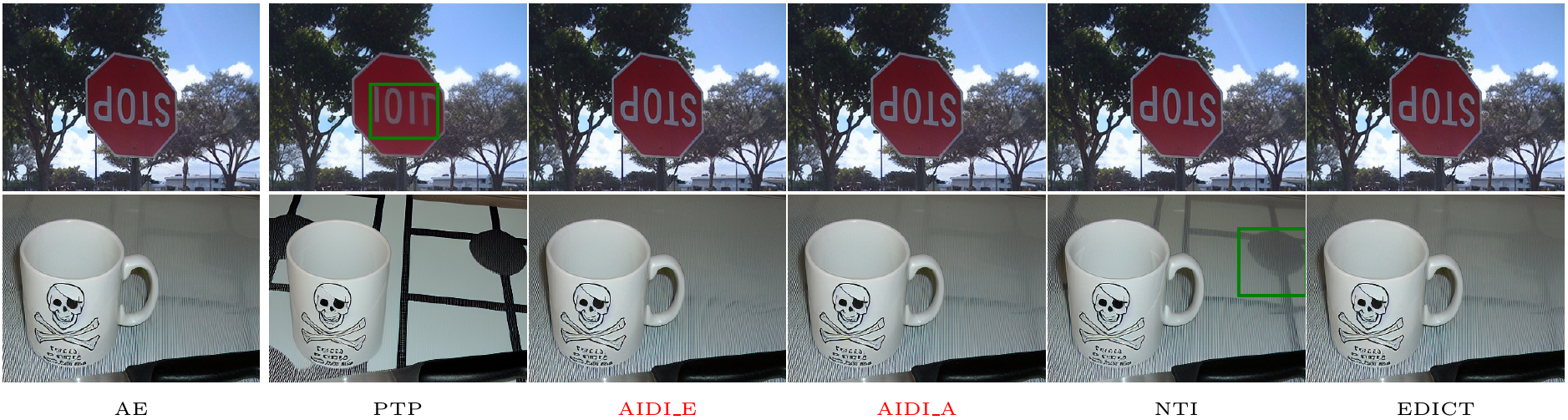}
 \end{center}
 \vspace{-6pt}
 \caption{Visual examples of reconstruction accuracy test for various diffusion inversion methods.  The AE images on the left are decoded from Stable Diffusion without inversion and used as the reference for perfect reconstruction.  Selected local artifacts are highlighted with \textcolor{green}{$\Box$} for quick reference.}
 \vspace{-6pt}
 \label{fig:recv}
\end{figure*}

%-------------------------------------------------------------------------
\subsection{Stochastic Editing}
\noindent For image editing methods based on deterministic DDIM inversion methods, the DDIM sampling process is commonly set as deterministic to better control
editing effects.  Similar to the proposed blended guidance, the same soft mask is adopted to control the affected area of stochastic sampling.  The deterministic sampling
process described in Equation~\ref{eq:dn} is made stochastic as:
 \begin{equation}
\resizebox{.86\width}{!}{$q(z_{t-1}) \sim \mathcal{N}(\sqrt{\bar{\alpha}_{t-1}} z^t_0 + \sqrt{1-\bar{\alpha}_{t-1}-\eta \sigma^2_t \Tilde{M}_t} \epsilon_t, \eta \sigma^2_t \mathbf{\Tilde{M}}_t)$} \label{eq:se}
 \end{equation}
where $\sigma^2_t$ represents the noise injected for stochastic sampling, and $\mathbf{\Tilde{M}}_t$ is transformed from ${\Tilde{M}}_t$ as a diagonal matrix.  Here $\eta$ and ${\Tilde{M}}_t$ control the scale and range of stochastic editing respectively.  While large $\eta$ increases sampling diversity, it reduces the perceptual similarity with the original image which is often desired in cases such as replacing an object with another in a different domain.  Based on our experiments, a small $\eta$ can result in satisfying results without loosing perceptual similarity when  deterministic editing fails.
%As this requires multiple samples, an editing method with fast sampling capability like our AIDI based process, which is stable for as few as 10 steps, is highly desired.

%%%%%%%%%%%%%%%%%%%%%%%%%%%%%%%%%%%%%%%%%%%%%%%%%%%%%%%%%%%%%%%%%%%%%%%%%
\section{Experiments}
\label{sec:exp}
\noindent All experiments conducted here are based on the released version
$\texttt{v1.4}$ of Stable Diffusion using NVIDIA A100 GPU card.  For image reconstruction test, 5000 test images from COCO dataset~\cite{lin_eccv_2014} are used.
For image editing, the test set of AFHQ~\cite{choi_cvpr_2020} is used primarily for its high image quality and because it enables comparisons with GAN inversion-based editing (we denote it as HS-SCLIP, combining HyperStyle~\cite{alaluf_cvpr_2022} and StyleCLIP~\cite{ding_arxiv_2021}).  Other than HS-SCLIP, we perform comprehensive comparisons in both reconstruction and editing quality with all related text-based real image editing techniques which use DDIM inversion, including PTP~\cite{hertz_arxiv_2022}, NTI~\cite{mokady_arxiv_2022} and EDICT~\cite{wallace_arxiv_2022}. NTI's main innovation is the null-text inversion, so in image editing tests it uses PTP editing techniques after the inversion.
%We also adopted one quantitative test of EDICT~\cite{} using 5 classes of test images from ImageNet~\cite{}, where the editing is applied to the background.
%Considering the downsampling used in Stable Diffusion, all images are resized to nearest ones that could be divided by 32 without a remainder.
For all visual examples, the default inversion and editing steps are 20 unless specified otherwise.
  
\begin{figure*}[t]
 \begin{center}
     \includegraphics[width=\linewidth]{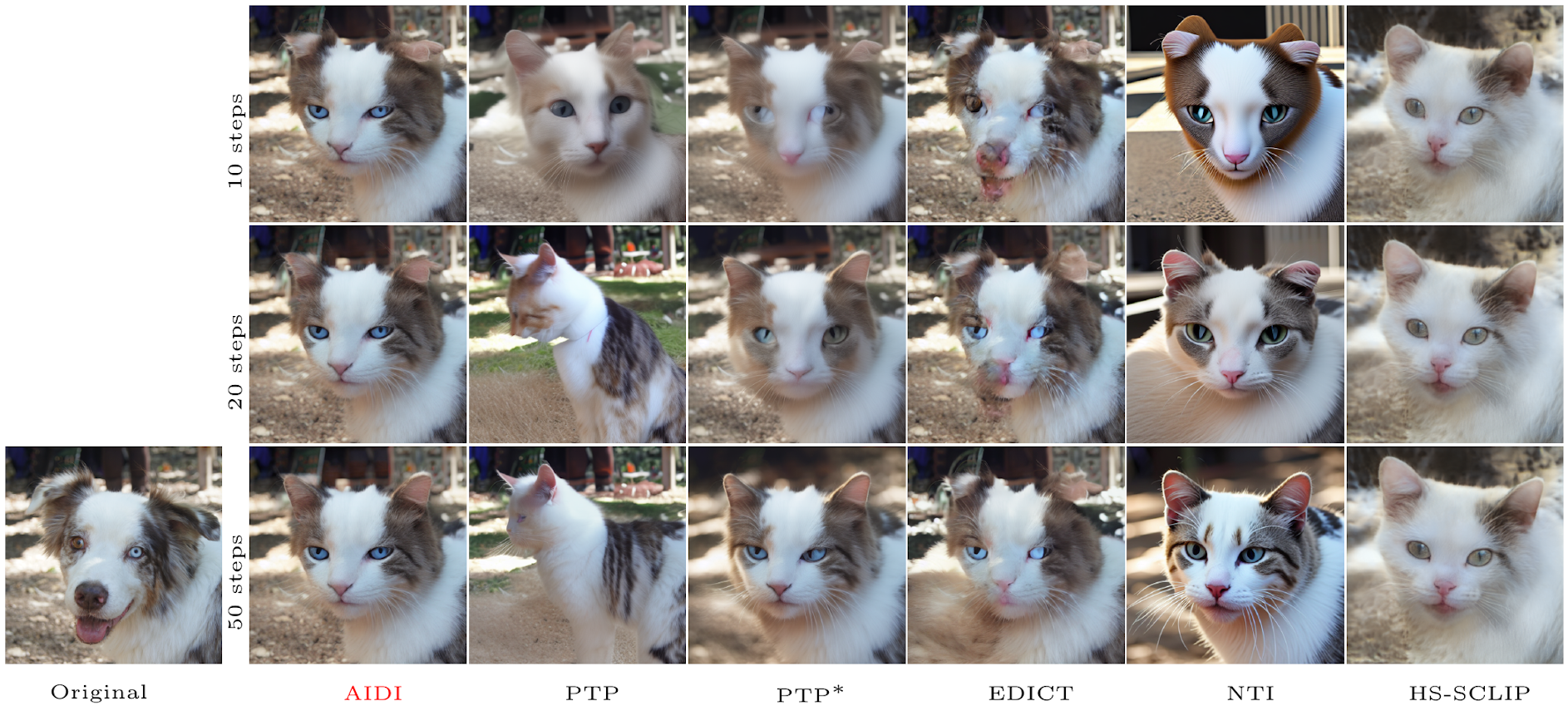}
 \end{center}
 \vspace{-6pt}
 \caption{Visual examples for dog-to-cat editing test using AFHQ test set.  Results from of different model are organized horizontally and results from different settings, 10/20/50 editing steps for all diffusion-based models or 3 hyperparameter settings for HS-SCLIP, are organized vertically.}
 \vspace{0pt}
 \label{fig:d2ca}
\end{figure*}

%-------------------------------------------------------------------------
\subsection{Image Editing}

\noindent Visual examples from Fig.~\ref{fig:edit} showcase the diverse capabilities of our proposed real image editing process.
Using a pair of text prompts as before and after-editing image captions, it can either perform
local editing, like replacing an object or background, or global editing, like style transfer.
As an example, for the \textit{zebra} image, we are able to change the background to a \textit{forest}
while keeping the unedited area perceptually unchanged.
For partial editing within one object, it can change the posture of the \textit{zebra} from
\textit{standing} to \textit{walking} in addition to editing the background, or
change the expression of the \textit{dog} to \textit{happy}.
It is also able to do a text-based style transfer, converting a photo to a \textit{watercolor painting}.
While similar range of editing capabilities have been demonstrated before, we are the first to demonstrate effective editing using as few as 10 editing steps, a result enabled by the improved accuracy of AIDI and by the proposed blended guidance, which focuses editing effects to specific relevant areas.
Note that without an exact binary mask, the local edits from these prompt-pair based processes would not be be able to keep certain areas intact.
Consider for example the editing task of changing the background of \textit{a castle in the mountains} to \textit{forest}: the castle itself remains largely the same but some greenery is blended within its structure, increasing the overall coherence of the resulting image.
This is arguably preferable to a hard blending with the background.
The lack of a binary mask can sometimes give rise to hallucinated artifacts, such as the lines around the mouth area in the \textit{dog wearing sunglasses} instance.

%-------------------------------------------------------------------------
\subsection{Reconstruction Accuracy}

\begin{figure}[h!]
 \begin{center}
     \includegraphics[width=\linewidth]{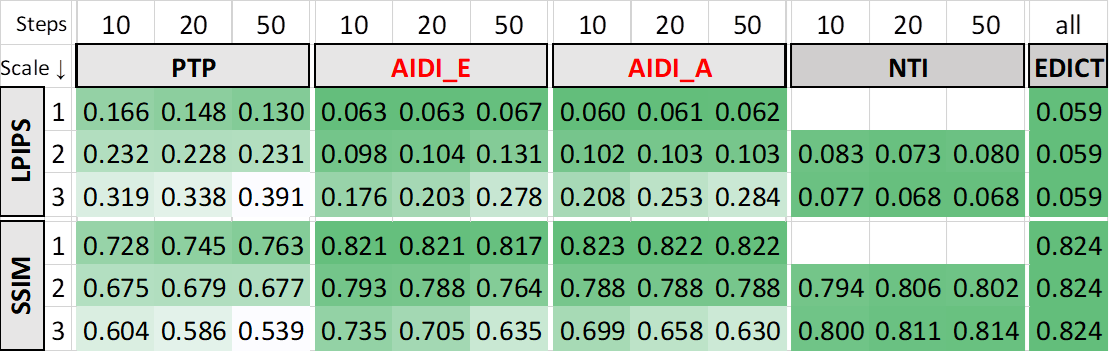}
 \end{center}
 \vspace{-6pt}
 \caption{Reconstruction accuracy test for different diffusion inversion methods using COCO test set.  Results from combinations of different inversion steps and various classifier-free guidance scales are included.  For the two assessed perceptual metrics, higher SSIM and lower LPIPS are preferred, both color-coded towards green color.}
 \vspace{-6pt}
 \label{fig:rec}
\end{figure}

\noindent We conducted a reconstruction test using the COCO test set with diverse contents using their default captions as text prompts where the quantitative results are illustrated as a graphic table, Fig.~\ref{fig:rec}.  Compared to the simple inversion
used in PTP, results from both our proposed AIDI variants are shown to significantly improve the accuracy across different classifier-free guidance scales and different diffusion steps.
For both AIDI\_E and AIDI\_A, they are able to achieve near-exact inversion when there is no classifier-free guidance.  While the accuracy drops when the guidance scale
increases, this does not impact our proposed real image editing process since a low guidance scale like 1 is used in inversion.   
We confirm that EDICT is able to maintain exact inversion with as few as 10 iteration steps, although it was only tested in the 50 steps' regime at publication.
As for NTI, it is only applicable to a guidance scale $> 1$.
Despite its accuracy increasing along with the guidance scale, it is worse, even at an empirically optimal guidance scale of 3,  than our proposed inversion and reconstruction with a guidance scale of 1.
For visual examples in Fig~\ref{fig:recv}, both PTP and NTI's results contain noticeable errors.
Our AIDI\_A result is as flawless as the EDICT one.
Despite the AIDI\_E result missing some minor detail, we find its differences from AIDI\_A negligible from both a quantitative and qualitative perspective.
Unless otherwise specified, all image editing tests are conducted using AIDI\_E.

%-------------------------------------------------------------------------
\subsection{Quantitative Assessment}
\noindent EDICT is the first one in diffusion inversion based image editing methods to report quantitative analysis for image editing.  We have conducted similar quantitative assessments using the AFHQ dataset. Our rationale for choosing AFHQ is that the high quality close-up images do make visual artifacts more noticeable. Another advantage is that for a dog-to-cat image test, there is a real cat image set to serve
as a photorealistic reference to evaluate editing results, an arguably better approach than using a CLIP score that relies on the indirect text reference.  This also enables comparison with GAN inversion-based methods, since a pre-trained StyleGAN model is available for this dataset.
Using the prompt pair $\textit{high quality photo of a dog} \rightarrow \textit{high quality photo of a cat}$, the task is to replace the dog in the input images with a cat which
not only looks realistic, but perceptually similar to the original dog in characteristics like color and pose, while maintaining the background unedited.  To evaluate the consistency in color, pose and backgrounds, the LPIPS metric is evaluated between the edited and original dog image.  To evaluate the editing effects, the FID score in reference to the AFHQ training set is used over CLIP score.
Results for other metrics, including CLIP, are included as supplementary materials.
\begin{figure}[t]
 \begin{center}
     \includegraphics[width=\linewidth]{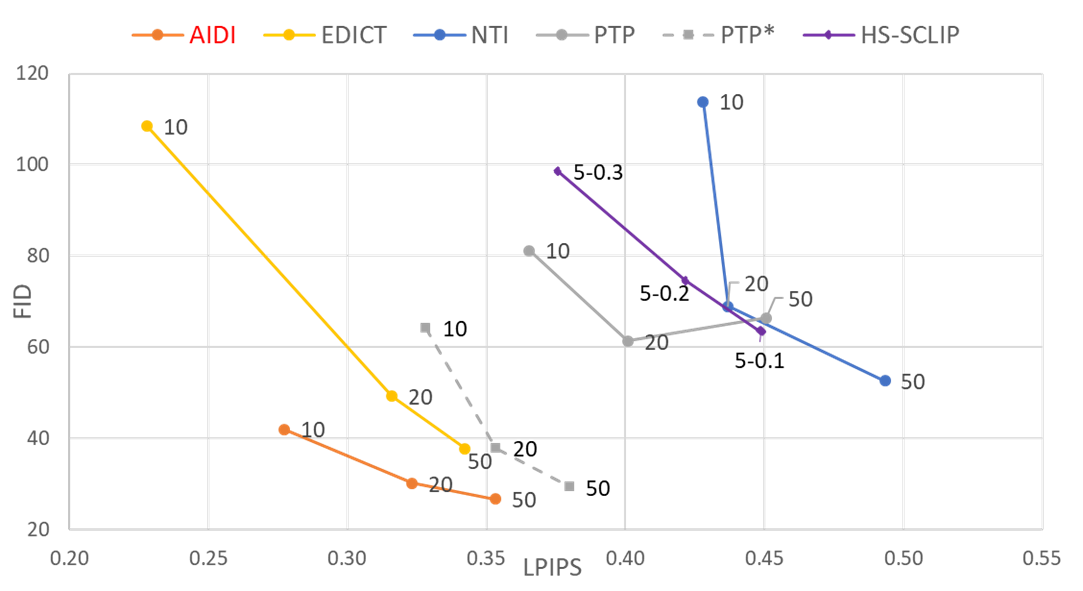}
 \end{center}
 \vspace{-6pt}
 \caption{Quantitative assessments for dog-to-cat swap test using AFHQ test set.  The data labels represent editing steps for all methods except for HS-SCLIP where they are hyperparameters.}
 \vspace{0pt}
 \label{fig:d2c}
\end{figure}

As shown in Fig.~\ref{fig:d2c}, our results are the best overall for any number of editing steps.  Our result of 20 editing steps is better than the next best EDICT at 50 steps.
In the extreme case of 10 steps, our method is still relatively effective while all other methods have significant regressions in quality.
For PTP, in addition to the original setting where the same guidance scale is used for both inversion and editing, a PTP$^*$ version is also tested where the inversion
is conducted without guidance, and it has significantly improved performance comparing to PTP.  Similar improvement has also been reported in EDICT, denoted as the difference between UC PTP and C PTP.  This dual-scale setting is similar to our blended guidance so the performance
gap between PTP$^*$ and ours is mainly caused by the increased inversion accuracy from our AIDI.  For HS-CLIP, which is based on HyperStyle inversion and StyleCLIP editing, three
sets of hyperparameters are included to illustrate similar trade-off between high perceptual similarity comparing to the source image and high editing quality in reference to the target domain.
To out knowledge, this is the first direct quantitative comparison between editing methods using diffusion inversion and GAN inversion respectively.

The results from NTI, PTP editing with null-text inversion, are worse than PTP baseline even with increased inversion stability.  From the visual examples in Fig.~\ref{fig:d2ca}, they seem to be able to generate a high quality cat image.  However, these images are often too smooth and the lack of photorealistic details
are causing degraded performance in both FID and LPIPS.
One possible cause is that the learned null-text embedding deviates from the original one too much so it is out-of-distribution for the pre-trained diffusion model.
Another consideration is that quantitative assessments like this require a fixed setting of hyperparameters so any method relying on fine-tuning of hyperparameters would not perform well on average.

The robustness of our methods for reduced number of inversion steps comparing to others is also noticeable in Fig.~\ref{fig:d2ca}.  For our method, although the 50-step result has better sharpness and more realistic ears, the 10-step version is of acceptable quality without any glaring artifacts, whereas all competing approaches are subject to
significant degradations at 10-steps.  In the case of EDICT, there are obvious artifacts, dog-like mouth area, for both 10-step and 20-step results.
Note that for this experiment, we used similar grid-search strategy
to find optimal settings for our method, PTP, PTP$^*$ and NTI, as we adopted attention map injection techniques from PTP.  For EDICT, its own default optimal settings are used.  For fair comparison, we didn't change hyperparameters for different editing steps, except for the fixed-point iteration steps in AIDI where 11, 6 and 5 are used for 10/20/50 editing steps respectively.

%-------------------------------------------------------------------------
\subsection{Stochastic Editing}
\begin{figure}[ht!]
 \begin{center}
     \includegraphics[width=\linewidth]{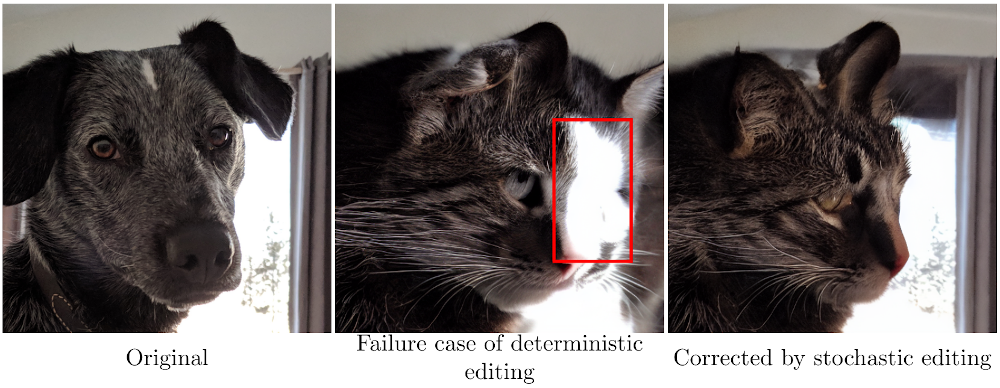}
     \includegraphics[width=\linewidth]{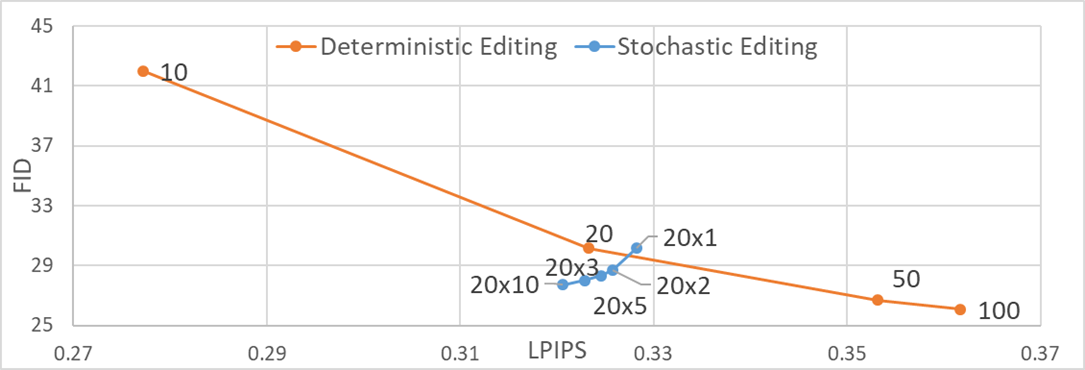}
 \end{center}
 \vspace{-6pt}
 \caption{Top: visual examples of stochastic editing recovers from failure case of deterministic editing; Bottom: quantitative comparison between deterministic and stochastic editing.}
 \vspace{0pt}
 \label{fig:pick}
\end{figure}

\noindent When using deterministic DDIM inversion, image editing through the DDIM sampling process must also be set deterministic in order to keep the fidelity of unedited areas.  As a result, there is no direct remedy for failed editing results.  As shown in the middle image in Fig.~\ref{fig:pick}, the original over-exposed background window was incorrectly edited as part of the cat face.  Using our proposed stochastic editing, it is possible to improve editing results when such failure case happens.
Different from other stochastic sampling practices which uses a larger $\eta$ for increased sampling diversity, we used a small
 $\eta = 0.1$ for stochastic generation of the same editing task and resulted in a properly edited result as shown on the right.  We conducted a quantitative analysis using the dog-to-cat test to measure this benefit.  As shown in the chart in Fig.~\ref{fig:pick}, $20\!\times\!n$ represents average performance when $n$ stochastic editings were repeated for each image to select the best one based on a combined rank of low FID and LPIPS.  It is shown that the stochastic editing, when tested only once, is expected to have an equivalent FID score as the deterministic one but worse in LPIPS.  With increased $n$, the selected result is expected to further improve in both FID and LPIPS.

%%%%%%%%%%%%%%%%%%%%%%%%%%%%%%%%%%%%%%%%%%%%%%%%%%%%%%%%%%%%%%%%%%%%%%%%%
\section{Conclusions}
\noindent In this work, we have proposed \emph{AIDI}, an accelerated iterative diffusion inversion method capable of improving image reconstruction and editing accuracy while
significantly improving the trade-offs between computational cost and quality. While our proposed real image editing is effective without auxiliary masks or sketches to specify editable area and content, it relies on the cross-attention maps between
the image and prompt texts for general guidance, similar to other text guided image editing work using diffusion models.  In addition to logically set prompt pairs, the spatial resolution of these attention maps are very coarse.  Detailed control of editable area remains a subject of future work.

It is also noted that while we don't use large classifier-free guidance scales for inversion with introduction of blended guidance, reducing the difference in guidance scales between inversion and editing is still useful to reduce potential artifacts caused by blended guidance.  While our AIDI can improve inversion stability for large guidance scales as well, it is not sufficient for reliable inversion yet and this is also of future research interests.

\noindent\textbf{Ethical considerations:} we acknowledge that any image editing technique such as the one presented in this paper are unavoidably encumbered by ethical concerns, and will inherit any bias present in the training data from the underlying backbone.  Deployments of these systems should employ appropriate safeguards in regards to allowed prompts and guidances to prevent malicious and illegal uses.

% While all the experiments we presented in this work are not conducted on images with inappropriate contents
% or of privacy concerns, we understand that many text-to-image diffusion models trained on a huge amount of data are not immune from generating undesirable
% results.  While practitioners adopting image editing techniques like ours should not use them for malicious purposes, they should also adopt additional
% safety measures in production settings to prevent generated inappropriate materials, even rare, from being releasing to the public.

{\small
\bibliographystyle{ieee_fullname}
\bibliography{mybib_2302}
}

\end{document}